%% file: 0_main.tex
\documentclass[final]{cvpr}

\usepackage{times}
\usepackage{epsfig}
\usepackage{graphicx}
\usepackage{amsmath}
\usepackage{amssymb}
\usepackage{arydshln} 
\usepackage{algorithm}      
\usepackage[noend]{algpseudocode} 
\usepackage{multirow}
\usepackage[dvipsnames]{xcolor}
\usepackage{color}
\usepackage{textcomp}
\usepackage{gensymb}
\usepackage{enumitem,amssymb}
\newlist{todolist}{itemize}{2}
\setlist[todolist]{label=$\square$}
\usepackage{pifont}
\newcommand{\cmark}{\ding{51}}%
\newcommand{\done}{\rlap{$\square$}{\raisebox{2pt}{\large\hspace{1pt}\cmark}}%
\hspace{-2.5pt}}

\newcommand{\workname}{ScaleNAS\xspace}
\newcommand{\netname}{ScaleNet\xspace}
\newcommand{\supernet}{SuperScaleNet\xspace}

\usepackage[pagebackref=true,breaklinks=true,colorlinks,bookmarks=false]{hyperref}
 
\usepackage{soul}

\begin{document}

\title{\workname: One-Shot Learning of Scale-Aware Representations for \\ Visual Recognition}

\author{Hsin-Pai Cheng$^{1,2}$\thanks{equal contributions.}\xspace\xspace, 
Feng Liang$^{1}$\footnotemark[1]\xspace\xspace, Meng Li$^2$, Bowen Cheng$^3$, Feng Yan$^4$, \\
Hai Li$^1$, Vikas Chandra$^2$ and Yiran Chen$^1$\\
$^1$Duke University, $^2$Facebook Inc, $^3$University of Illinois at Urbana-Champaign, $^4$
University of Nevada, Reno\\
}

\maketitle

\begin{abstract}
Scale variance among different sizes of body parts and objects is a challenging problem for visual recognition tasks. 
Existing works usually design a dedicated backbone or apply Neural architecture Search(NAS) for each task to tackle this challenge.
However, existing works impose significant limitations on the design or search space.
To solve these problems, we present \workname, a one-shot learning method for exploring scale-aware representations. 
\workname solves multiple tasks at a time by searching multi-scale feature aggregation. 
\workname adopts a flexible search space that allows an arbitrary number of blocks and cross-scale feature fusions.
To cope with the high search cost incurred by the flexible space, \workname employs one-shot learning for multi-scale supernet driven by grouped sampling and evolutionary search. 
Without further retraining, \netname can be directly deployed for different visual recognition tasks with superior performance.
We use \workname to create high-resolution models for two different tasks,
\netname-P for human pose estimation and \netname-S for semantic segmentation. 
\netname-P and \netname-S outperform existing manually crafted and NAS-based methods in both tasks. 
When applying \netname-P to bottom-up human pose estimation, it surpasses the state-of-the-art HigherHRNet. In particular, \netname-P4 achieves 71.6\% AP on COCO test-dev, achieving new state-of-the-art result. 

\end{abstract}
\input{1_intro}

\input{2_related_work}
\input{3_ScaleNAS}
\input{4_expt}
\input{5_discussion}

\input{6_conclusion}
{\small
\bibliographystyle{ieee_fullname}
\bibliography{egbib}
}
\input{7_supp}

\end{document}

%% file: 1_intro.tex
\section{Introduction}

\begin{figure}
    \centering
    \includegraphics[width=1.0\linewidth]{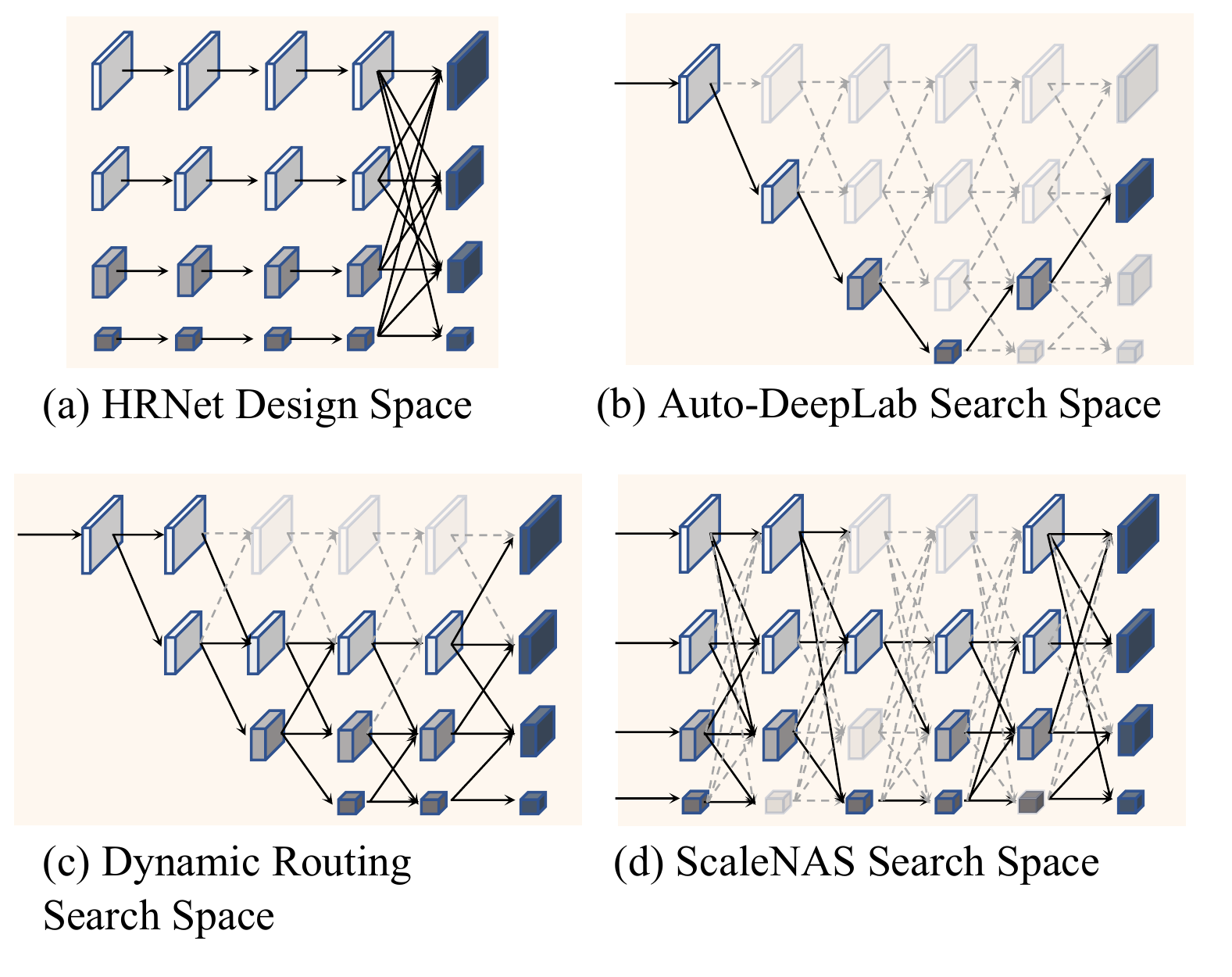}
    \caption{
    Search space comparison. (a) HRNet uses fully connected multi-scale feature fusion at the end of every four blocks. (b,c) Auto-DeepLab and dynamic routing allows feature fusion connection for each neighboring feature map and find the best architectures with single-path and multi-path, respectively. (d) We propose a more flexible feature fusion that allows crossing to remote feature maps to maximize multi-scale aggregation.
    }\label{fig:comp_search_space}
\end{figure}

\begin{figure}[t]
    \centering
    \includegraphics[width=1.0\linewidth]{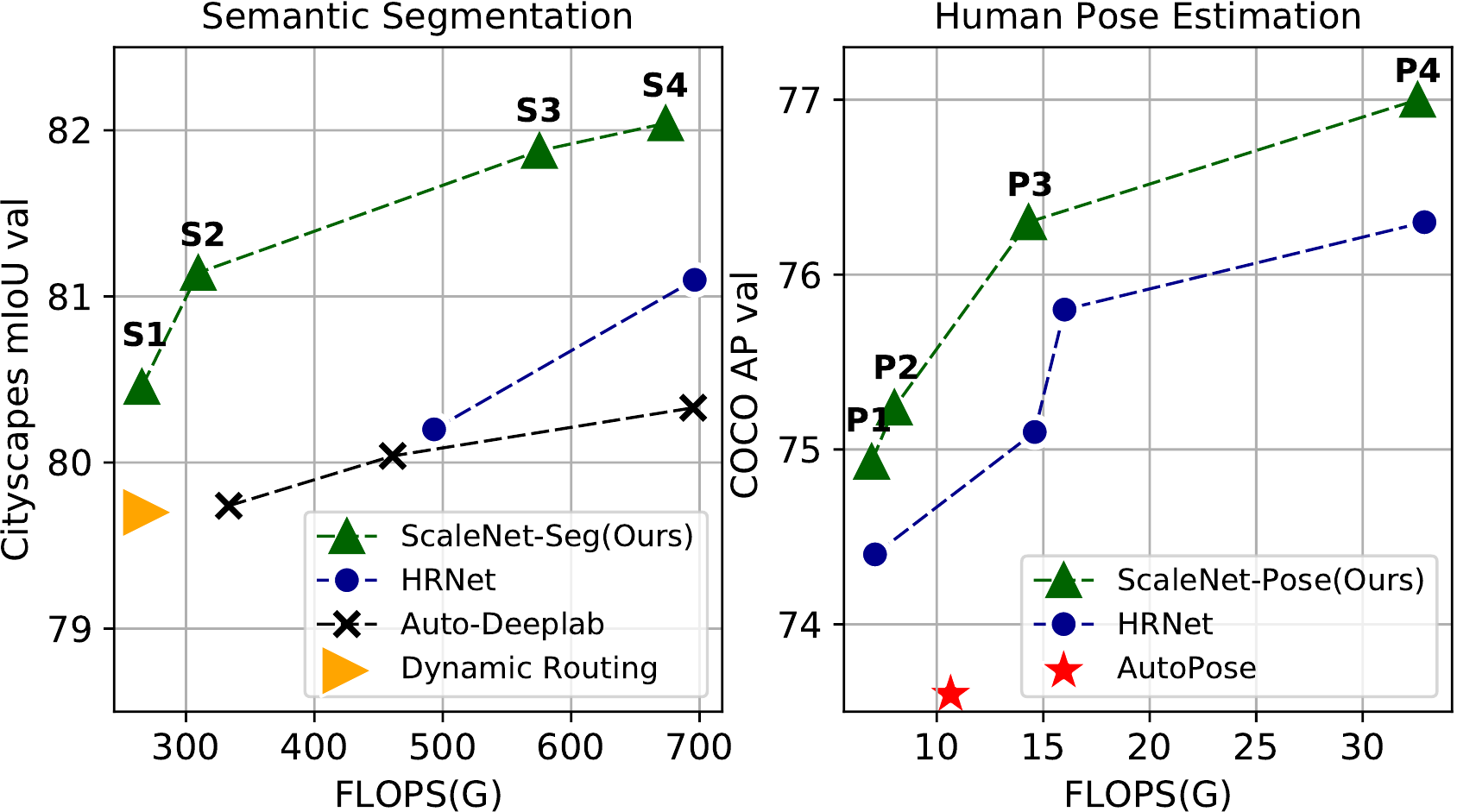}
    \caption{The trade-off between computation cost (GFLOPs) and model performance. Left: semantic segmentation mIoU on Cityscapes val. Right: human pose estimation AP on COCO val. Our \netname outperforms HRNet and NAS-based methods.} 
    \label{fig:pareto_seg_pose}
\end{figure}

Deep-learned representation can be generally categorized into low-resolution representation and high-resolution representation.
Low-resolution representation is typically used in classification tasks while
high-resolution representation is essential for visual recognition tasks such as semantic segmentation and human pose estimation. 
We focus on the high-resolution representation in this paper.
There are three important yet challenging considerations when designing high-resolution representation:
1) scale variance from different sizes of objects and scenes; 2) precise and informative feature maps are critical;
3) different deployment platforms have different model size requirements.

\textbf{Challenge of scale variance:}
take semantic segmentation as an example, the variance of object size induces difficulty for
pixel-level dense prediction, and thus scale representation is critical.
In human pose estimation, 
it is challenging to localize human anatomical keypoints when there is a large scale variance in the scene such as tiny and large persons, the large difference in joint distance. 
We address this challenge by proposing a new multi-scale search space.

\textbf{Challenge of high-resolution representation:}
to design high-resolution representations, earlier efforts recover high-resolution representations from low-resolution outputs, e.g., Hourglass~\cite{newell2016stacked}, SegNet~\cite{badrinarayanan2017segnet}, U-Net~\cite{ronneberger2015u}. Recent works focus on maintaining high-resolution representation through the whole network and aggregating different scale of representation from parallel paths. E.g., HRNet~\cite{sun2019deep, wang2020deep} and its variants use such multi-scale high-resolution networks to achieve state-of-the-art results on human pose estimation. 
However, multi-scale neural architectures usually have a large design space to explore and are prone to design redundancies. 
Our study reveals that when different scales of representation have different depths, the performance can be greatly improved. 

\textbf{Challenge of deriving a wide spectrum of models:}
previous methods can only derive one architecture at a time. To obtain different sizes of model, further retraining is required for each candidate architecture,  
e.g., it takes $O(N)$ training time to derive $N$ models.
We propose one-shot based searching method to lower the training cost to $O(1)$.

To address the above challenges, we propose \workname, a one-shot based searching method to explore scale-aware neural architectures.
We tackle the scale variance challenge by proposing \textit{multi-scale aggregation search space} to explore multi-scale aggregation and network depth for high-resolution representation, see Figure~\ref{fig:comp_search_space}.
Under this search space, we propose a one-shot based searching method to discover multiple  architectures simultaneously. 

We name these elite architectures \netname, which can be directly deployed without retraining while performing as well as stand-alone models. As demonstrated in Figure~\ref{fig:pareto_seg_pose},
\netname outperforms manually crafted and NAS-based models on semantic segmentation and human pose estimation. In semantic segmentation, \netname surpasses HRNet, Auto-DeepLab, and dynamic routing by 1.3\% - 1.7\% mIoU on CityScapes dataset with less computation cost. In human pose estimation, our  ScaleNet-P4 obtains 71.6\% AP on COCO \textit{test-dev2017}, achieving a new state-of-the-art result on multi-person pose estimation leaderboard.
We further study the patterns of \netname by analyzing the trade-offs between convolution blocks and feature fusion for multi-scale neural architecture design.

%% file: 2_related_work.tex
\section{Related Work}
\noindent\textbf{High-Resolution Neural Architectures.}
Designing high-resolution neural architectures is important yet challenging. For example, 
Hourglass~\cite{newell2016stacked} uses a high-to-low followed by a low-to-high architecture to achieve high-resolution.
SimpleBaseline~\cite{xiao2018simple} uses transposed convolution layers to generate high-resolution representations. 
To prevent information loss from the encoding-decoding process, HRNet~\cite{wang2020deep} proposed to keep high-resolution features at all times and use multi-scale feature fusion at the end of every four residual blocks to merge information from different scales. While this method has proven successful in many vision tasks, such manual design has many redundant operations and is not optimal as we demonstrated in Figure~\ref{fig:teaser_seg}. 

\noindent\textbf{NAS for High-resolution Architectures.}
Recent  works proposed automating the design of neural architectures for semantic segmentation and human pose estimation. For example, 
Auto-DeepLab and dynamic routing~\cite{li2020learning} are proposed to search architectures for semantic segmentation models. 
PoseNAS~\cite{bao2020pose}, AutoPose~\cite{gong2020autopose} and PNFS~\cite{yang2019pose} are proposed to search architectures for human pose estimation.

However, the above NAS-based methods can only craft architectures for one task at a time. Given different platforms have different deployment constraints, previous works used adjustable factors (e.g., channel width) to scale neural architecture to different sizes~\cite{wang2020deep}. After scaling, it is required to retrain each of the scaled architecture. Therefore, to get $N$ models, it requires $O(N)$ training time. 
Since this procedure is costly and time consuming, we proposed a one-shot based searching method that can derive $N$ models in $O(1)$ time without any retraining.

\noindent\textbf{One-shot Neural Architecture Search.}
One-shot NAS aims at searching a large neural architecture and sharing weights to different sub-networks~\cite{pham2018efficient, bender2018understanding, liu2018darts, cai2018proxylessnas, guo2020single}. While these weights have to adapt to different sub-networks, one-shot NAS suffers from low accuracy. Recent works applied various techniques to conquer the low accuracy problem of supernet. For example, BigNAS~\cite{yu2020bignas} transforms the problem of training supernet to training a big single-stage model and applies sandwich rule to guarantee the performance for each sub-network. OFA~\cite{cai2020once} proposed to use progressive shrinkage together with knowledge distillation to train a one-shot supernet. However, these methods are mainly designed for single-path neural architecture on relatively simple task (e.g., ImageNet classification). 
Directly applying existing one-shot training methods to multi-scale architecture yeilds sub-optimal performance.
In this paper, we solve this problem by our proposed \textit{grouped sampling method} to effectively explore a wide spectrum of sub-networks and further search elite architectures with our evolutionary method.

%% file: 3_ScaleNAS.tex
\section{\workname}
In this section, we first identify the search space problems in existing works by performing a search space exploration.
Then we introduce our proposed \textit{multi-scale aggregation search space}. Based on this search space, we train a \textit{one-shot based \supernet} that contains a wide spectrum of architecture candidates. Finally, we employ \textit{multi-scale topology evolution} to derive elite \netname based on a trained \supernet. 

\subsection{Search Space Exploration}
Existing works that achieve state-of-the-art results on semantic segmentation and human pose estimation impose limitations on the design space.
As shown in Figure~\ref{fig:comp_search_space},
HRNet supports cross-scale feature fusions, but it uses four residual blocks in every scale of branchs. Such regular design results in redundancy and misses optimization opportunities as the depth for each branch can be altered to improve performance. 

Auto-DeepLab~\cite{liu2019auto} includes multiple scale options in their search space to search a single-path neural architecture for semantic segmentation. 
Dynamic routing~\cite{li2020learning} reused the search space form Auto-DeepLab to search a multi-path neural architecture to achieve improved performance on semantic segmentation. 
Although Auto-DeepLab and dynamic routing search the connections between different scales of feature maps, the search space for fusion is limited to only neighboring scales.
Such limitation restricts the representation ability for feature maps and cross-scale feature fusion can provide better information gathering for each scale of feature maps. 

\begin{figure}[h!]
    \centering
    \includegraphics[width=0.8\linewidth]{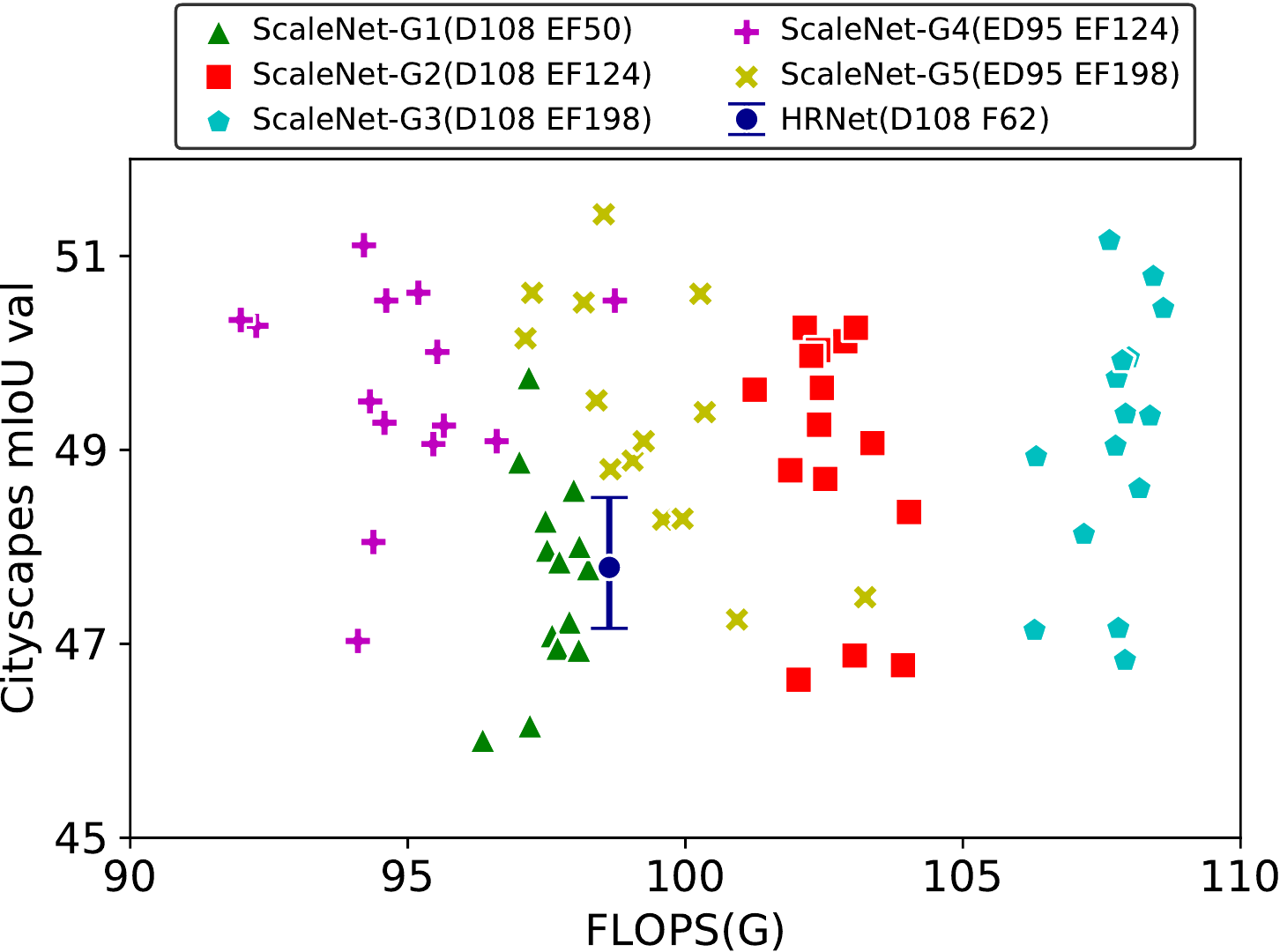}
    \caption{Search space exploration. We train HRNet five times and record the mean and variance of their accuracy. HRNet has 108 resblocks (denoted as `D') and 62 fusions(denoted as `F'). We randomly sample 5 groups of \netname with different resblocks and fusions based on our search space (Figure~\ref{fig:comp_search_space}(d)). `ED' and `EF' represents the expectation of blocks and fusions, respectively. 
    }
    \label{fig:teaser_seg}
\end{figure}

To illustrate our search space, we compare the proposed search space with existing works in Figure~\ref{fig:comp_search_space}.
Different from existing works, we provide flexible depth (number of residual blocks) for each scale of branch. In addition, we allow feature fusion cross to any other scale of branches.
We use this search space to randomly sample neural architectures as \netname-G series (Figure~\ref{fig:teaser_seg}) and train them on Cityscapes.
Original HRNet has 108 residual blocks~\cite{he2016deep} and 62 feature fusions. 
Residual block is composed of two $3\times3$ convolutions. Multi-scale fusion includes downsampling and upsampling. For downsampling, we use strided $3 \times 3$ convolution with stride 2. For upsampling, we use bilinear upsampling followed by a $1\times1$  convolution for aligning the number of channels~\cite{wang2020deep}.
We create ScaleNet-G1 by using the same number of blocks as HRNet while 
using 12 less feature fusions with our proposed search space, we observe that there are some ScaleNet-G1 models perform better than HRNet while having less number of fusions. Therefore, the feature fusion position of HRNet may not be optimal as shown in Figure~\ref{fig:teaser_seg}.
Based on \netname -G1, we create \netname -G2 and \netname -G3 by increasing the number of fusions to the expectation of 124 and 198, respectively. We notice that more feature fusions comes with a higher mIoU and inevitable comes with higher FLOPs. To study the redundancy of number of blocks, we create G4 and G5 by decreasing number of blocks while keeping higher number of fusions. We observe that the mean accuracy of \netname -G4, G5 are still higher than the original HRNet setting. Based on this observation, we envision that we can use neural architecture search to explore the trade-offs and relationships between blocks and fusion connection. 

\begin{figure*}[ht]
    \centering
    \includegraphics[width=1\linewidth]{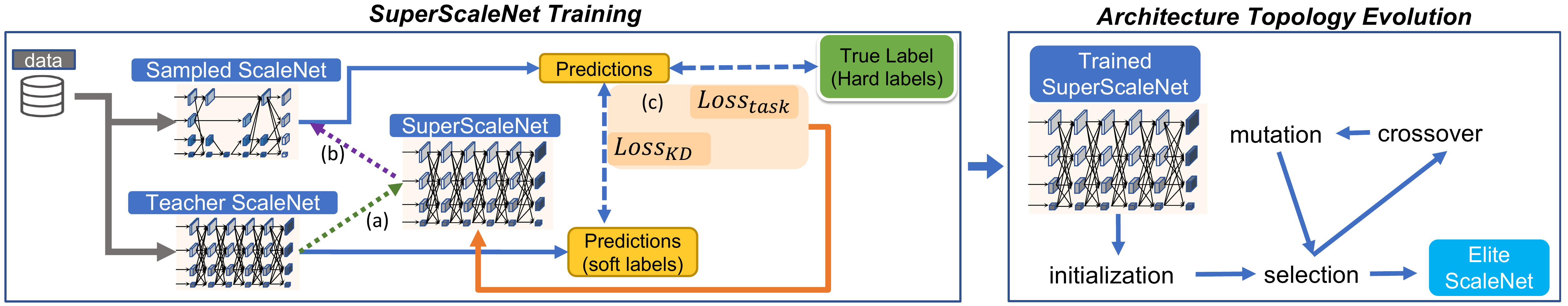}
    \caption{Workflow of \workname.  \workname train a SuperScaleNet in the proposed search space and uses our proposed evolutionary method to explore elite architectures based on the trained SuperScaleNet.    (a) Before training starts, we initialize SuperScaleNet by the teacher model. (b) During each iteration, we sample ScaleNet from the SuperScaleNet. (c) We use the task loss from true labels and the knowledge distillation (KD) loss from soft labels given by teacher to update \supernet.}
    \label{fig:workflow}
\end{figure*}

\begin{figure*}[ht]
    \centering
    \includegraphics[width=1\linewidth]{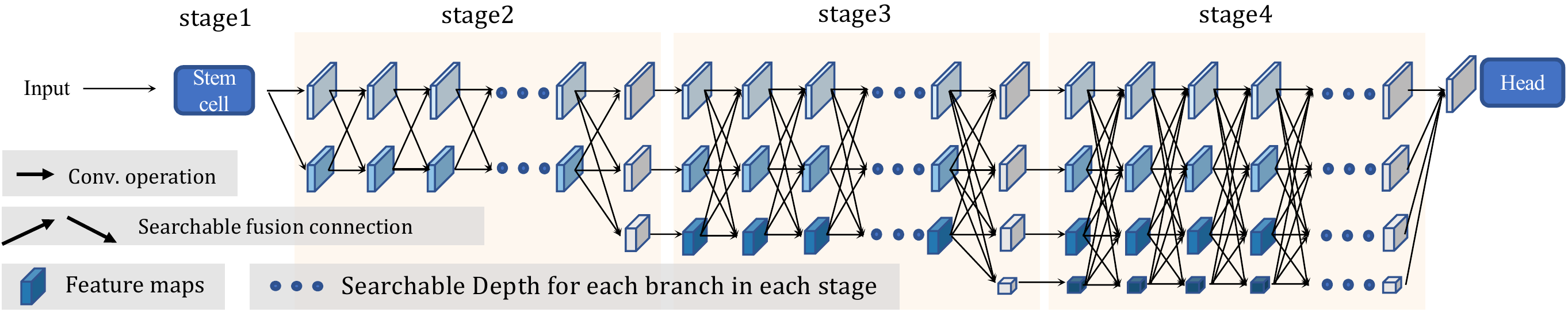}
    \caption{Search space overview of \workname. Our search space inherit the spirit of HRNet that has few stages. \workname adopts a flexible search space with arbitrary number of blocks and cross-scale feature fusions.
  }
    \label{fig:search_space}
\end{figure*}

\subsection{Multi-scale Aggregation Search Space}
\label{sec:search_space_design}

Our goal is to design multi-scale neural architectures that can be adapted to multiple visual recognition tasks without retraining. Instead of searching a single model at a time, we aim at discovering a wide spectrum models that have different computation cost for different deployment scenarios. 

We employ a stage-based search space design, which is inspired by the state-of-the-art architecture HRNet that can be adapted to multiple tasks.
The architecture starts from a stem cell as the first stage and it is composed of two stride-2 3$\times$3 convolutions. There are four stages in our search space. After the first stage, we gradually add one more high-to-low branch for the following stages, i.e., stage2, stage3, and stage4 maintains two-resolution, three-resolution, and four-resolution branches respectively.

Based on the initial experiments shown in Figure~\ref{fig:teaser_seg}, we observe that fusion percentages has positive correlation with accuracy. 
Thus four blocks per branch proposed by HRNet is not always optimal. 
We introduce two Controlling factors to form our search space: \
\begin{enumerate}
    \item Branch depth ($d$). Instead of searching the overall depth for the entire network, we allow a more flexible search space that can search depth for each branch in individual independent stage module. For simplicity, the depth of each branch is chosen from \{d1, d2, d3, d4\}. In this paper, we currently set it at \{2, 3, 4, 5\}
    \item Fusion percentage ($f$). Here the fusion percentage is defined as the probability of the out-degree fusion for each feature map. E.g., a feature map with fusion percentage of 100\% means this feature map connects to every other scales of fusion in its current depth.

\end{enumerate}

By relaxing the cross-scale feature fusion and enlarging the branch blocks, we have roughly 7$\times$10$^{72}$  different neural network architectures in our search space.

\subsection{Training one-shot \supernet}
In this section, we explain how to design a one-shot training for multi-scale aggregation search space. 
Recall that cross-scale feature fusion can boost up performance (see Figure~\ref{fig:teaser_seg}).
Here we propose to jointly train \supernet together with a teacher model that has sufficient feature fusion connection.  

Figure~\ref{fig:workflow} depicts the workflow of \supernet training. 
First, according to the search space proposed in Figure~\ref{fig:search_space}, we build the teacher model with full number of blocks and fully activated feature fusions. 
Visual recognition tasks usually rely on ImageNet pretrained to stabilize training~\cite{wang2020deep,ren2015faster}, we pretrain the teacher model on ImageNet until converge. 
Then, we initialize \supernet with the weights from the teacher model. During each training iteration, we sample a sub-network from \supernet, pass one batch of training data to both sampled sub-network and teacher model. 
Next, we calculate the task loss using the true label, and knowledge distillation (KD) loss using the soft label given by teacher model.
Finally, we update the supernet based on the combination of both task loss and KD loss. 
Our training objective is fomulated as follow:
\begin{equation}
\min_{W_s} \sum_{arch_i} (\mathcal{L}_{task}(P_{arch_i}, y)+ \alpha \cdot MSE(P_{arch_i}, P_t)) .
\end{equation}

Our main goal is to optimize the weights of \supernet ${W_s}$
with the combination of true label loss and soft label loss.
Here $P_{arch_i}$ stands for the prediction for each sampled architecture, $y$ is the true label. $P_t$ stands for the prediction from teacher model. 
We use MSE to calculate the loss between sub-network prediction and teacher network prediction. The KD ratio ($\alpha$) is set to 1.

\subsection{Grouped Sampling}
\label{sec:grouped_sampling}
To train supernet, sampling plays a crucial role. 
Sandwich rule~\cite{yu2019universally,yu2020bignas} was proposed  to train supernet, where  the smallest model, the largest model, and 2 randomly sampled models are trained in every iteration.
However, we observe that sandwich rule cannot guarantee to explore a wide spectrum of neural architectures. Therefore, we propose grouped sampling by dividing the whole search space into different sub-groups. 

Given a depth choice of \{d1, d2, d3, d4\}, we group the depth choice to \{[d1, d2], [d2, d3], [d3, d4]\} (e.g., \{[2, 3], [3, 4], [4, 5]\}), the fusion percentage is selected from \{f1, f2, f3\}  (e.g., \{0.2, 0.5, 0.8\}). In combination, we have a total of 9 (3$\times$3) sub-groups.
Compared with sandwich rule, grouped sampling is more suitable for multi-scale aggregation search space. Empirical justification is detailed in Section~\ref{sec:ablation_grouped_sampling}.

It is worth noting that unlike in OFA~\cite{cai2020once} or BigNAS~\cite{yu2020bignas}, where 4 sub-networks are sampled and their gradients are aggregated in each update step, our group sampling only get one sub-network in each iteration. Therefore, the cost of training a \supernet using group sampling is very low, e.g., equivalent to a standard model training. 

\subsection{Multi-scale Architecture Topology Evolution}
\label{sec:topology_evolution}

To explore sub-network from a trained supernet, existing one-shot based methods use coarse-to-fine selection, predictor-based method, etc. 
However, these methods are mainly designed for single-path neural architecture. 
In comparison, our search space includes multiple path in each stage and each stage has different number of scales.
Therefore, we propose \textit{multi-scale architecture topology evolution}, which provides more reasonable and controllable crossover and mutation over candidate architectures.  

As described in Algorithm~\ref{alg:evo} and Figure~\ref{fig:workflow}, our topology evolution include the following four phases.
\begin{itemize}
    \item \textbf{Step1 Initialization.} We uniformly sample $n_0$ sub-networks and record their architectures, accuracies, and FLOPs as a set $\mathbf{D}$. $n_0$ equals to 1,000 in our experiments.
    \item \textbf{Step2 Selection.} We select top $k$ models on the Pareto front of cost/accuracy trade-off curve in $\mathbf{D}$ as candidate group $\mathbf{C}$. For each subnetwork in $\mathbf{C}$, we do crossover and mutation to obtain next-generation offsprings. $k$ is set to 100 in our experiments.
    \item \textbf{Step3 Crossover.} Since different stage module may have different number of branches, our crossover is inner-stage crossover. For each sub-network $\mathbf{arch_c}$ in $\mathbf{C}$, we allow a probability $p_{c}$ to swap stage module settings with another randomly selected sub-network. There are 8 stage modules (1,4,3 for stage 2, 3, 4, respectively) and $p_{c}$ is set to 0.25. Thus, each sub-network is expected to have 2 modules been replaced.
    \item \textbf{Step4 Mutation.} After crossover, we do random mutation to switch on and off the fusion connections in every stage module with probability $p_{m}$. $p_{m}$ is set to 0.5. These $K$ offspring models along with their corresponding accuraices and FLOPs are recorded as set $\mathbf{M}$. Then we update $\mathbf{D}= \mathbf{D}\cup\mathbf{M}$. 
    We continue to Step2 until we have $N$ sub-networks in $\mathbf{D}$. Here we set $N$ as 2,000.
\end{itemize}

\renewcommand{\algorithmicrequire}{\textbf{Input:}}
\renewcommand{\algorithmicensure}{\textbf{Output:}}
\begin{algorithm}
    \caption{Multi-Scale Architecture Evolution}
    \label{alg:evo}
    \begin{algorithmic}[1]
        \Require{
        Search space $S$, 
        Trained \supernet, initial population size $n_0$,
        number of offspring $k$, crossover probability $p_{c}$, mutation probability $p_{m}$}, 
        number of final elite architectures $N$.
        \Ensure{$N$ Elite \netname}
        \State Sample $n_0$ sub-networks to obtain initial population
        $\mathbf{D}=\{arch_d, d=1,2,3, \dots, n_0\}$
        \While {len$(\mathbf{D}) < N$} 
        \State Select top $k$ models on the Pareto front as candidate group  $\mathbf{C}=\{arch_c, c=1,2,3, \dots, k\}$
        \For {every sub-networks $\mathbf{arch_c}$ in $\mathbf{C}$}
        \State crossover and mutation under probability $p_{c}$ and $p_{m}$ sequentially, generate offspring
        \EndFor
        \State Gather $k$ offspring models as set $\mathbf{M_k} = \{arch_m, m = 1,2,\dots, k\}$
        \State update $\mathbf{D}= \mathbf{D}\cup\mathbf{M}$
        \EndWhile

    \end{algorithmic}
\end{algorithm}

%% file: 4_expt.tex
\section{Experiments} \label{sec:experiment}

In this section, we evaluate \workname by searching neural architectures for semantic segmentation and human pose estimation. First we train \supernet on semantic segmentation with Ctiyscapes dataset~\cite{cordts2016cityscapes} and derive \netname-S using \workname. Then we apply the same searching routine on top-down human pose estimation framework with COCO dataset~\cite{lin2014microsoft} to derive \netname-P. 
In order to evaluate the generalizability of \netname, we apply \netname-P to HigherHRNet framework for bottom-up human pose estimation.
Finally, we conduct ablation studies for \workname.

\noindent\textbf{Training setup.}
To stabilize training, we first train the teacher model with full depths and fusions on ImageNet-1k~\cite{deng2009imagenet} dataset. Following the training procedure in ~\cite{wang2020deep}, we train teacher model for 100 epochs. More training details can be found in the supplementary material.

\subsection{Semantic Segmentation}
\label{sec:semantic_seg}
\begin{table*}
\centering
\caption{Semantic segmentation results on Cityscapes \textit{val} (single scale and no flipping). The GFLOPs is calculated on the input size 1024 × 2048. `D-X' equals to `Dilated-X'. For existing segmentation NAS works, the total cost grows linear to the number of deployment scenarios $N$, while the cost of our \workname remains constant.} 
\setlength{\tabcolsep}{4pt}
\small
\begin{tabular}{c||c|c|c|c||c|c|c} 
\hline
\multirow{2}{*}{Method}   & \multirow{2}{*}{Backbone} &  \multirow{2}{*}{\#Params}  & \multirow{2}{*}{GFLOPs} & mIoU  & Searching Cost & Training Cost & Total Cost($N$=40)        \\ 
   &  &   &  & (\%)  & (GPU hours) & (GPU hours) & (GPU hours)      \\ 
\hline \hline
DeepLabv3~\cite{chen2017rethinking} & D-ResNet-101 &  58.0M &  1778.73 &  78.5 & - & 50$N$ & - \\
DeepLabv3+~\cite{chen2018encoder}& D-Xception-71 &  43.5M &  1444.63 &  79.6 & - & - & - \\
PSPNet~\cite{zhao2017pyramid} & D-ResNet-101 &  65.9M &  2017.63 &  79.7 & - & 100$N$ & - \\
Auto-DeepLab~\cite{liu2019auto} & Searched-F20-ASPP & - & 333.3 & 79.7 & 72$N$ & 250$N$ & 12.9$k$ \\
Dynamic Routing~\cite{li2020learning} & Layer33-PSP & - & 270.0 & 79.7 & 180$N$ & ~\textbf{0} & 7.2$k$  \\
\workname(Ours) & \netname-S1   & ~\textbf{25.3M}    & ~\textbf{265.5}  & 80.5 & 200 &  400 & 600 \\
\workname(Ours) & \netname-S2   & 28.5M    & 309.5  & ~\textbf{81.1} & ~\textbf{200} &  400 & ~\textbf{600} \\
\hdashline
Auto-DeepLab~\cite{liu2019auto} & Searched-F48-ASPP & - & 695.0 & 80.3 & 72$N$ & 350$N$ & 16.9$k$ \\
HRNet~\cite{wang2020deep}       & HRNet-W48       & ~\textbf{65.8M}  & 696.2  & 81.1  & - &  260$N$  & -\\
\workname(Ours) & \netname-S4   & 67.5M                     & ~\textbf{673.6}  & ~\textbf{82.0} &  ~\textbf{300} & ~\textbf{600} & ~\textbf{900} \\

\hline
\end{tabular}
\label{tab:segmentation}
\end{table*}
\noindent\textbf{Cityscapes.} The Cityscapes~\cite{cordts2016cityscapes} is a widely used dataset for semantic segmentation tasks, which contains 5,000 high quality pixel-level finely annotated scene images. The dataset is divided into 2975/500/1525 images for training, validation, and testing, respectively. There are 30 classes, and 19 classes among them are used for evaluation. The mean of class-wise intersection over union (\textit{mIoU}) is adopted as our evaluation metric. 

\noindent\textbf{Implementation details.} We first obtain the teacher model following the same training protocol in~\cite{zhao2018psanet, wang2020deep}. 
The teacher model is trained for 484 epochs with the batch size of 24. 
After the teacher model is trained, we use our grouped sampling technique (Section~\ref{sec:grouped_sampling}) to further fine tune the \supernet-Seg to support smaller sub-networks. 
More training details can be found in the supplementary material.

\noindent\textbf{Segmentation results.} 
Table~\ref{tab:segmentation} reports the comparison between \workname and existing manual/NAS methods on semantic segmentation. Comparing with NAS (Auto-Deeplab and dynamic routing), \workname is much more efficient for multiple deployment scenarios. 
E.g., when there are 40 deployment scenarios, the total cost of \workname s 12× fewer than dynamic routing and 19× fewer than Auto-Deeplab, respectively. Without additional retraining, \netname-S1 outperforms the dynamic routing Layer33-PSP by a 0.8\% margin under the similar cost. When comparing with manually designed HRNet-W48 or Searched-F48-ASPP, \netname-S4 improves the mIoU to 82.0\%, surpassing HRNet and Auto-Deeplab by 0.9\% and 1.7\% respectively.

\subsection{Human Pose Estimation}
For human pose estimation, we first search \netname-P on top-down human pose estimation task using COCO~\cite{lin2014microsoft}. Then we reuse the searched \netname-Pose on MPII~\cite{andriluka20142d} and bottom-up pose estimation tasks.

\noindent\textbf{COCO.}
We train \supernet-Pose on COCO \textit{train2017} dataset (57K images and 150K person instances) and evaluate it on COCO \textit{val2017}. To evaluate object keypoints, we use Object Keypoint Similairty (OKS). We break down the performance on  different OKS: $AP_{50}$ and $AP_{75}$. We also report the performance on different sizes of object. $AP_{M}$ and $AP_{L}$ stands for AP of medium object and large object, respectively.

\noindent\textbf{MPII.}
The MPII Human Pose dataset~\cite{andriluka20142d} consists real-world full-body pose and annotations. There are around 25K images with 40K subjects, where 12K subjects are used for testing and the remaining subjects are used for training. We use the PCKh (head-normalized probability of correct keypoint) score as our evaluation metric, following~\cite{xiao2018simple, sun2019deep}

\subsubsection{Top-down Methods}

\noindent\textbf{Implementation details.} We use the same workflow as semantic segmentation. Following the settings of~\cite{sun2019deep, wang2020deep}, the teacher model and \supernet-Pose are trained for 210 epochs. More training details can be found in the supplementary material.

\noindent\textbf{Top-down results.} Table~\ref{tab:pose_top_down} summarizes the results of top-down methods on COCO \textit{val2017} and MPII \textit{val}, compared with other state-of-the-art methods. Under 256$\times$192 input resolution, our \netname-P2 outperforms manually designed SimpleBaseline~\cite{xiao2018simple}(+3.2\%) and NAS based AutoPose~\cite{gong2020autopose}(+1.6\%) by a large margin. In addition, \netname-P2 is comparable with the strong HRNet~\cite{wang2020deep} baseline but with only 56\% parameters and 55\% FLOPs. With 384$\times$288 input resolution, our \netname-P3 achieves 76.3\% AP on COCO \textit{val2017}, outperforming PoseNFS-3~\cite{yang2019pose} by 3.3\% AP with less computation cost. \netname-P3 has the same accuracy as HRNet-W48 but uses only 42\% parameters and 43\% FLOPs. \netname-P4 obtains 77.0\% AP, surpassing its strong HRNet counterpart by 0.7\% AP. For MPII, \netname-P1 performs better on every body parts comparing with SimpleBaseline and HRNet.

\begin{table*}
\centering
\caption{Top-down human pose estimation results.
}
\small
\begin{tabular}{c||c|c|c|c|c|c|c|c|c|c} 
\hline
\multicolumn{11}{c}{\footnotesize{Comparison on COCO \textit{val2017}. AutoPose* reports results without ImageNet pretraining.}} \\
\hline
Method & Backbone & Input size  & \#Params & GFLOPs & AP & AP$_{50}$  & AP$_{75}$  & AP$_{M}$  & AP$_{L}$  & AR \\ 
\hline 
SimpleBaseline~\cite{xiao2018simple}  & ResNet-152       & \multirow{4}{*}{256$\times$192} & 68.6M    & 15.7   & 72.0 & 89.3      & 79.8      & 68.7     & 78.9     & 77.8 \\
AutoPose~\cite{gong2020autopose}   & AutoPose*       &  & -    & 10.65   & 73.6 & 90.6      & 80.1      & 69.8     & 79.7     & 78.1 \\
HRNet~\cite{wang2020deep}           & HRNet-W48        &  & 63.6M    & 14.6   & 75.1 & ~\textbf{90.6}      & 82.2      & 71.5     & 81.8     & 80.4 \\
\workname(Ours) & \netname-P2  &  & ~\textbf{35.6M}    & ~\textbf{8.0}    & ~\textbf{75.2} & 90.4      & ~\textbf{82.4}      & ~\textbf{71.6}     & ~\textbf{81.9}     & ~\textbf{80.4} \\
\hline
PNFS~\cite{yang2019pose}   & PoseNFS-3       & \multirow{5}{*}{384$\times$288} & -    & 14.8    & 73.0 & -      & -      & -     & -    &  \\
SimpleBaseline~\cite{xiao2018simple}  & ResNet-152       &  & 68.6M    & 35.6   & 74.3 & 89.6      & 81.1      & 70.5     & 79.7     & 79.7 \\
HRNet~\cite{wang2020deep}          & HRNet-W48        &   & 63.6M    & 32.9   & 76.3 & 90.8      & 82.9      & 72.3     & 83.4     & 81.2 \\
\workname(Ours) & \netname-P3  &  & ~\textbf{26.2M}    & ~\textbf{14.3}   & 76.3 & 90.7      & 82.9      & 72.5     & 83.3     & 81.3  \\
\workname(Ours) & \netname-P4 &  & 64.3M    & 32.6   & ~\textbf{77.0} & ~\textbf{90.9}      & ~\textbf{83.6}      & ~\textbf{73.0}     & ~\textbf{84.2}     & ~\textbf{81.8}  \\
\hline
\hline
\multicolumn{11}{c}{\footnotesize{Comparison on MPII \textit{val}. The GFLOPs is calculated on the input size 256 × 256. We reuse the searched \netname-P and apply it to MPII dataset.}} \\
\hline
Method  & Backbone & \#Params & GFLOPs  & mean &Head & Shoulder & Elbow &  Wrist & Hip & Knee     \\ 
\hline 
SimpleBaseline~\cite{xiao2018simple} & ResNet-152 & 68.6M & 20.9 & 88.5 & 96.4 & 95.3  & 89.0  & 83.2 & 88.4 & 84.0\\
HRNet~\cite{wang2020deep} & HRNet-W32 & 28.5M &  9.5 & 90.3 & 97.1 & 95.9 & 90.3 & 86.4 & 89.1 & 87.1\\
ScaleNAS (Ours) & \netname-P1 &  \textbf{28.5M} & \textbf{9.3} &\textbf{91.0} & \textbf{97.3}  & \textbf{96.5} & \textbf{91.5} & \textbf{87.3} & \textbf{90.0} & \textbf{87.5} \\

\hline
\end{tabular}
\label{tab:pose_top_down}
\end{table*}

\subsubsection{Bottom-up Methods}
We plug the elite architectures obtained from top-down human pose estimation into
state-of-the-art bottom-up human estimation framework HigherHRNet~\cite{cheng2020higherhrnet}. 

\noindent \textbf{Implementation details.}
We adopt the standard training procedure on COCO \textit{train2017} as in ~\cite{newell2017associative, cheng2020higherhrnet} and report results on COCO \textit{val2017} and \textit{test-dev2017}. The models are trained for 300 epochs. More training details can be found in the supplementary material.

\noindent \textbf{Bottom-up results.}
Tabel~\ref{tab:pose_bottom_up} reports the results of bottom-up methods on COCO \textit{val2017} and \textit{test-dev2017}. By utilizing our \netname-P as feature extractor, we boost the performance of bottom-up pose estimation. Our ScaleNet-P4 and ScaleNet-P1 outperform their counterparts by 0.7\% AP and 0.5\% AP on COCO \textit{val2017} ,respectively. In particular, our ScaleNet-P4 obtains 71.6\% AP on COCO \textit{test-dev2017} without using refinement or other post-processing techniques, achieving a new state-of-the-art result on multi-person pose estimation leaderboard.

\begin{table}
\centering
\caption{Bottom-up human pose estimation results.} 
\setlength{\tabcolsep}{2pt}
\footnotesize
\begin{tabular}{c||c|c|c|c|c} 

\hline
\multicolumn{6}{c}{Comparison on COCO \textit{val2017} w/o multi-scale test.} \\
\hline
\multirow{2}{*}{Method} & \multirow{2}{*}{Backbone} & Input  & \multirow{2}{*}{\#Params} & \multirow{2}{*}{GFLOPs} & \multirow{2}{*}{AP} \\
                  &   & size  &  &   &  \\
\hline
\multirow{4}{*}{HigherHRNet~\cite{cheng2020higherhrnet}}& HRNet-W32  & 512  & 28.6M & 47.9 & 67.1 \\
& \netname-P1(Ours) & 512 & 28.6M & 46.9 & 67.8  \\
& HRNet-W48  & 640  & 63.8M & 154.3 & 69.9 \\
& \netname-P4(Ours) & 640 & 64.4M & 141.5 & 70.4 \\
\hline \hline
\multicolumn{6}{c}{Comparison on COCO \textit{test-dev 2017} w/ multi-scale test.} \\
\hline
\multirow{2}{*}{Method} & \multirow{2}{*}{Backbone} & Input  & \multirow{2}{*}{\#Params} & \multirow{2}{*}{GFLOPs} & \multirow{2}{*}{AP} \\
                  &   & size  &  &   &  \\
\hline
Hourglass~\cite{newell2017associative} & Hourglass & 512 & 277.8M& 206.9 &63.0 \\
Hourglass w/ refine~\cite{newell2017associative} & Hourglass & 512 & 277.8M & 206.9 &65.5 \\
PersonLab~\cite{papandreou2018personlab}  & ResNet-152 & 1401 & 68.7M & 405.5 &68.7 \\
HigherHRNet~\cite{cheng2020higherhrnet}& HRNet-W48 & 640 & \textbf{63.8M} & 154.3 & 70.5 \\
ScaleNAS (ours) & \netname-P4(Ours) & 640 & 64.4M & \textbf{141.5} & \textbf{71.6} \\
\hline
\end{tabular}
\label{tab:pose_bottom_up}
\end{table}

\subsection{Ablation Study}
We perform ablation study on each proposed technique.
All results are conducted with \supernet-Seg-W32 on Cityscapes. For simplicity, we denote \supernet-Seg-W32 as supernet in this part.

\begin{figure}
    \centering
    \includegraphics[width=0.7\linewidth]{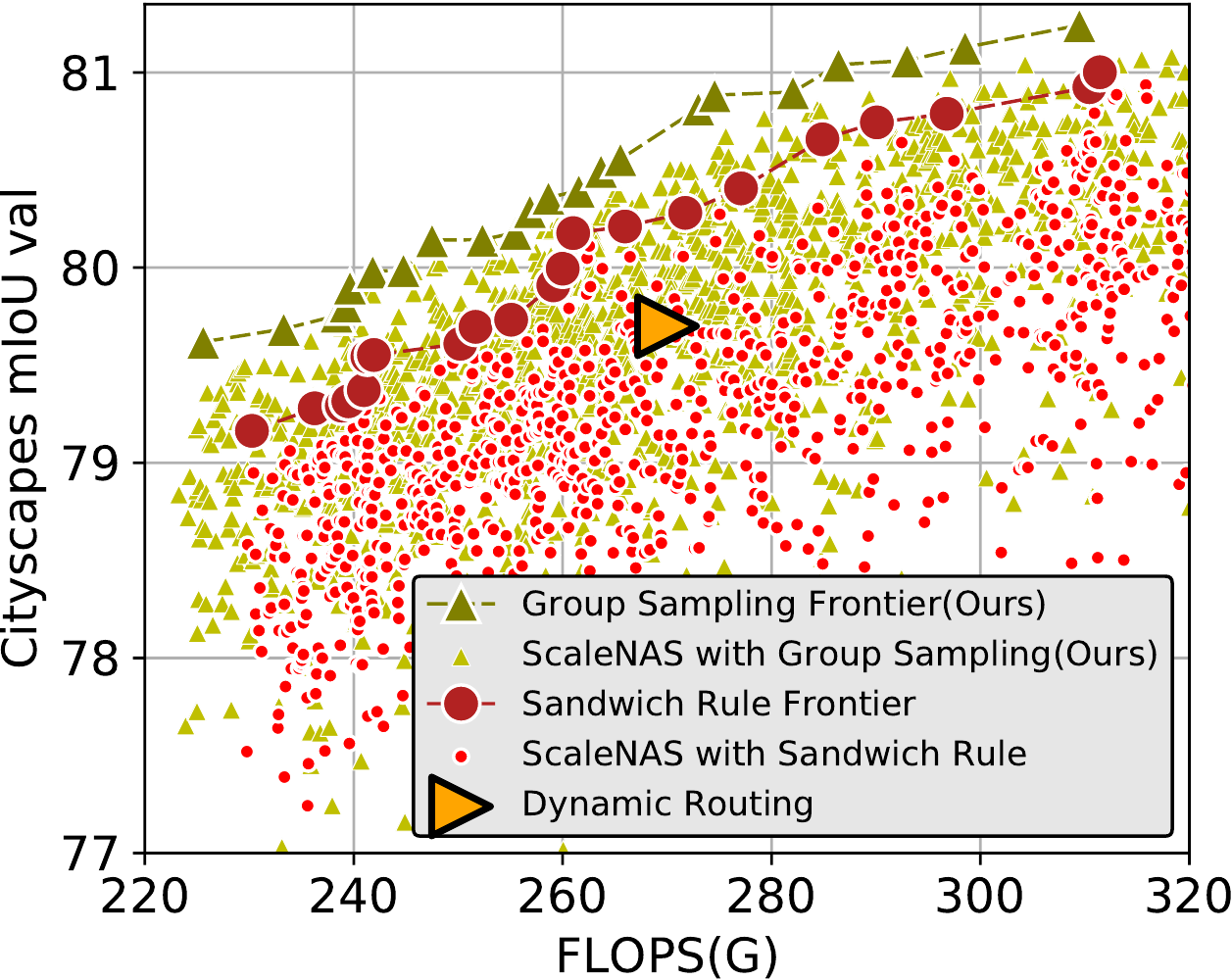}
    \caption{
    Ablation study of sampling techniques. The Pareto front of grouped sampling steadily higher than the Pareto front achieved by sandwich rule.
    }
    \label{fig:grouped_sample}
\end{figure}

\noindent\textbf{Impact of sampling.}\label{sec:ablation_grouped_sampling}
To study the impact of sampling technique, we train two supernets: one is based on our proposed method (Section~\ref{sec:grouped_sampling}), the other is based on state-of-the-art sampling method -- sandwich rule~\cite{yu2020bignas}. 
We derive the Pareto front from these two supernets based on our proposed evolutionary search. In Figure~\ref{fig:grouped_sample}, we show elite architectures and their corresponding accuracy from 220G to 320G. The Pareto front results suggest grouped sampling perform the best for multi-scale aggregation search space which has a wide spectrum of architectures.

\begin{figure}
    \centering
    \includegraphics[width=0.7\linewidth]{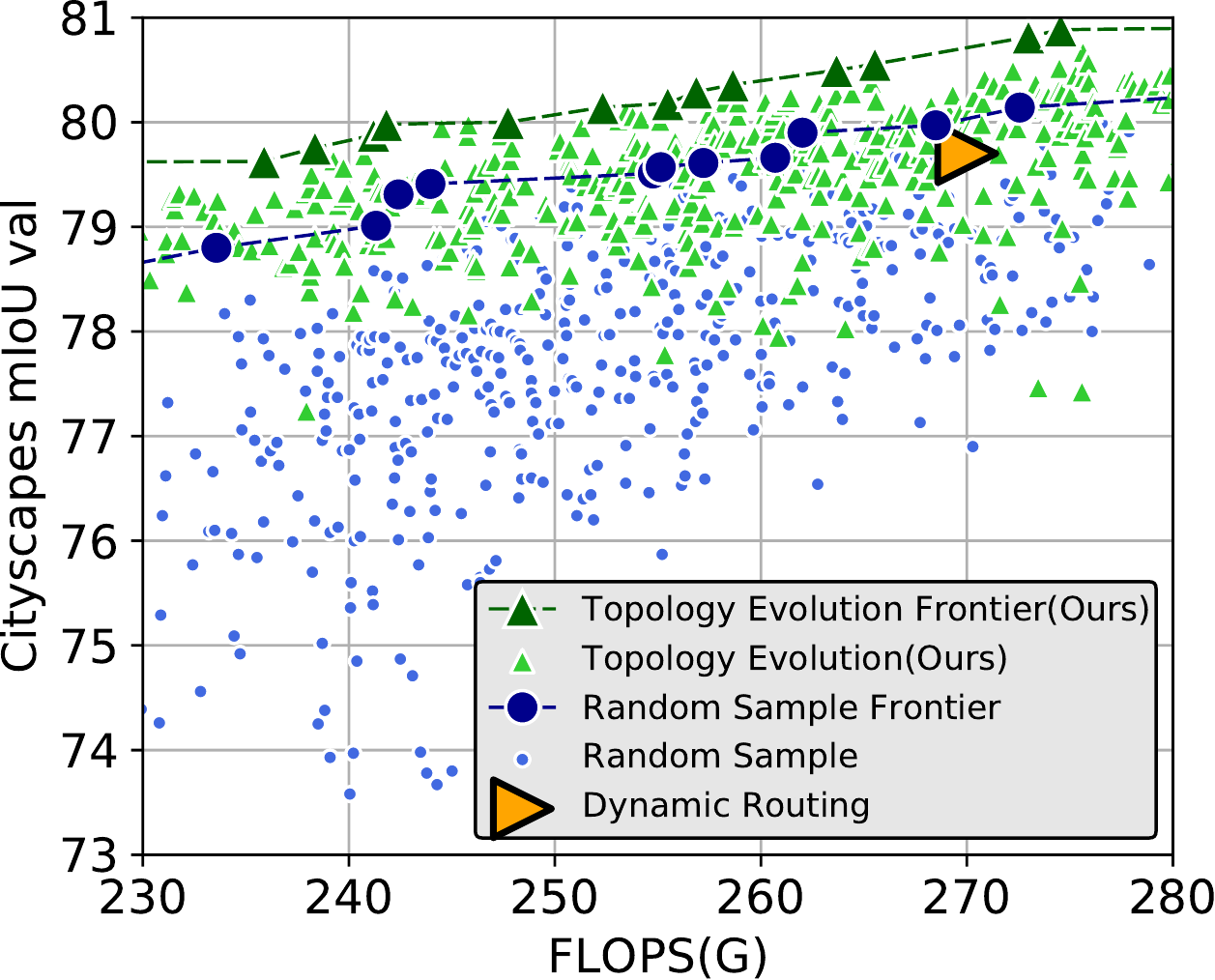}
    \caption{Ablation study of topology evolution. Dynamic routing is at around the Pareto curve of random sampling. The Pareto front of grouped sampling consistently higher than random sampling and dynamic routing.
    }
    \label{fig:abl_evo}
\end{figure}

\noindent\textbf{Impact of topology evolution.}
We demonstrate the performance contribution from searching method by comparing with random search. 
We use random sampling to sample the same amount of architectures as our evolutionary method and plot the Pareto front. As shown in Figure~\ref{fig:abl_evo}, under the same searching budget, our multi-scale topology evolution consistently achieves better performance on Pareto front, thanks to our inner-stage crossover and mutation techniques.

\label{sec:ablation_topology_evolution}

\noindent\textbf{Impact of knowledge distillation.}
We further study whether knowledge distillation (KD) plays an important role in accuracy gain.
Based on the same training procedure in Section~\ref{sec:semantic_seg}, we train \netname -S1 from ImageNet pretrained weights (stand-alone) with and without KD. We use MSE loss with KD ratio 1 and the teacher model pretrained on Cityscapes. As shown in Table~\ref{tab:ablation_standalone_training}, the accuracy of the stand-alone training is only slightly lower than directly taken from supernet. It suggests that KD is beneficial but not a dominant factor in the final accuracy.

\begin{table}
\centering
\caption{Ablation study of knowledge distillation (KD). Comparison with stand-alone training with and without KD. The performance mIoU(\%) is obtained on Cityscapes \textit{val}.} 
\small
\begin{tabular}{c||c|c|c} 
\hline
\multirow{2}{*}{Model} & from  & stand-alone  & stand-alone  \\
 & supernet & w/o KD  & w/ KD  \\
\hline
\netname-S1 & 80.5 & 80.2  &  80.4  \\
\hline
\end{tabular}
\label{tab:ablation_standalone_training}
\end{table}

\label{sec:standalone_training}

%% file: 5_discussion.tex
\section{Discussion}

\noindent\textbf{Crafted architectures.}
We demonstrate the crafted architectures for both semantic segmentation and human pose estimation in Figure~\ref{fig:found_arch}. We show the first module for each stage (full model description is in supplementary material). We observe that both architectures have various cross-scale feature fusion, which is important for handling large scale variance in these tasks. 
In addition, we observe later stages rely on heavier feature fusions while earlier stages have less feature fusions. 

\begin{figure}[t]
    \centering
    \includegraphics[width=0.9\linewidth]{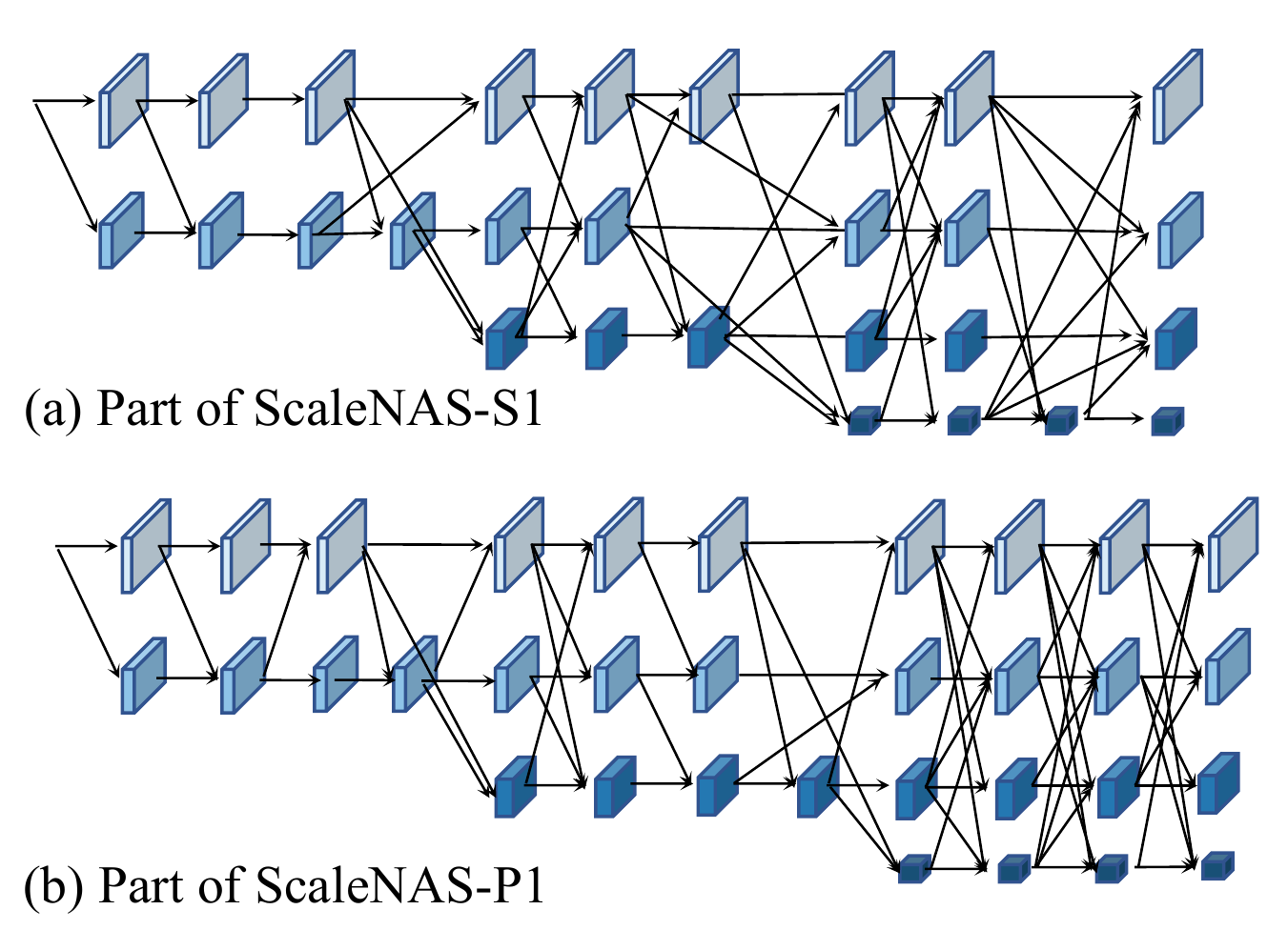}
    \caption{Architecture demonstration for ScaleNet-S1 and ScaleNet-P1. }
    \label{fig:found_arch}
\end{figure}

\noindent\textbf{Elite architecture pattern analysis.}
Different hardware platforms have different computation constraints. 
We analyze the deployability of elite architectures with different computation cost.
We record FLOPs, the number of fusions, and the number of blocks of Pareto front from  the 2000 elite ScaleNets collected by our evolutionary method. 

\begin{figure}[t]
    \centering
    \includegraphics[width=1.0\linewidth]{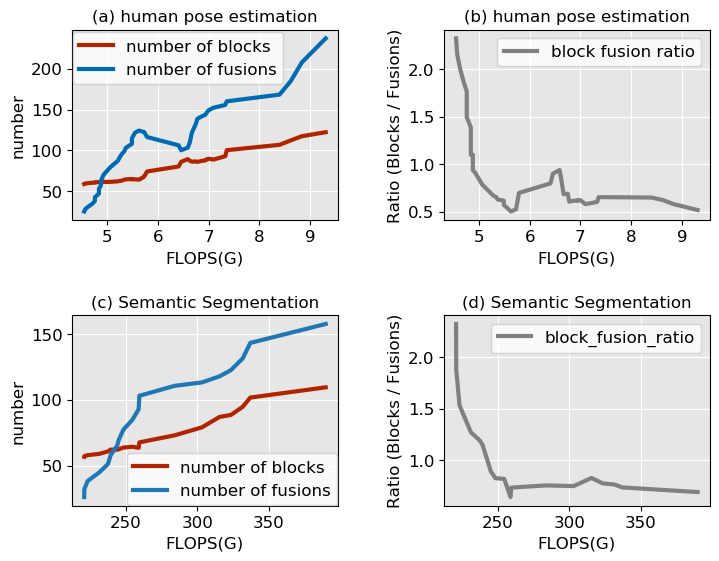}
    \caption{The network pattern of elite sub-networks. We show the relationship between number of blocks and fusions for the elite sub-networks.}
    \label{fig:fusion_analysis}
\end{figure}

In Figure~\ref{fig:fusion_analysis}(a)(c), we observe larger elite models have more fusions than blocks. 
In addition, the number of fusions increases faster than the number of blocks. To further analyze the relationship between number of fusions and number of blocks, we demonstrate block-fusion ratio in Figure~\ref{fig:fusion_analysis}(b)(d).
We observe that for the largest elite model, it requires two times more fusions than blocks. However, for small elite models, the number of fusions is only half of the number of blocks. 
This interesting observation provides important future design insights: 1) For edge devices, we should invest more computation cost on blocks than fusions. 2) To design larger models, it is preferable to invest computation cost on fusions over blocks.

%% file: 6_conclusion.tex
\section{Conclusion}

We present \workname, a one-shot learning method for scale-aware representations.
To the best of our knowledge, \workname is the first of its kind one-shot NAS method that considers scale variance for multiple vision recognition tasks. 
To efficiently search a wide spectrum of neural architectures for different vision tasks, we rest upon the following key ideas:
(i) A novel multi-scale feature aggregation search space that 
includes cross scale feature fusions and flexible depths. 
(ii) One-shot based training method driven by an efficient sampling technique to train multi-scale supernet.
(iii) Multi-scale architecture topology evolution to efficiently search elite neural architectures. 
All the above novel ideas coherently make \workname outperform existing hand crafted and NAS-based methods on semantic segmentation and human pose estimation. 

%% file: 7_supp.tex
\section{Supplementary Material}
This supplementary material provides more details of training \supernet on each task and also the extension of object detection task. 
For \textit{reproducibility}, \textbf{we provide full searching and training codes, as well as pretrained models}. Please refer to \texttt{README.md} to see detailed instructions.

\begin{table*}[t]
\centering
\caption{Object detection results on COCO $minival$ in Faster R-CNN~\cite{ren2015faster} and Mask R-CNN~\cite{he2017mask}. LS denotes learning rate scheduler. GFLOPs is calculated on the input size 800×1280. HRNet-w32* denotes our reimplementation.} 
\begin{tabular}{c|c|c|c|cccc|cccc} 
\hline
\multirow{2}{*}{Backbone}  & \multirow{2}{*}{LS} & \multirow{2}{*}{Params(M)} & \multirow{2}{*}{GFLOPs}   & \multicolumn{4}{|c|}{box}& \multicolumn{4}{c}{mask}\\
\cline{5-12}
&&& & AP   & AP$_S$  & AP$_M$  & AP$_{L}$  & AP  & AP$_{S}$ & AP$_{M}$ & AP$_{L}$   \\ 
\hline 
\multicolumn{11}{c}{Faster R-CNN~\cite{ren2015faster}} \\
\hline

HRNet-w32*    &  1$\times$ & 47.2  & 285.4 & 39.8 & 22.8 & 43.7 & 51.0 & / & /& /& / \\
\textbf{\netname-S2}  & 1$\times$ & 46.3  &  271.3 & 40.1  & 23.9  & 44.2  & 51.7  & / &/ & /& /     \\

\hline 
\multicolumn{11}{c}{Mask R-CNN~\cite{he2017mask}} \\
\hline
HRNet-w32*   & 1$\times$ &  49.9 &  353.9  &  40.8 & 23.8 & 44.5 & 52.4 & 36.4 & 19.5 & 39.7  & 48.9 \\
\textbf{\netname-S2}  &   1$\times$  & 49.0  &  339.8  & 40.9 & 24.4 & 44.6 & 52.5 & 36.5 & 19.7 & 40.0  & 49.0   \\
\hline
\end{tabular}

\label{tab:object_detection}
\end{table*}

\subsection{Details of Search Space Exploration}
In our main submission, we conduct initial search space exploration on semantic segmentation using Cityscapes. All models are trained from scratch for 48 epochs. Data augmentation strategies and other training protocols are the same as the teacher training part of Section~\ref{semantic_seg_training_details} in this supplementary material.

\subsection{Details of Training Teacher Model on ImageNet}

Following the instructions in~\cite{wang2020deep}, we use stochastic gradient descent (SGD) as the optimizer with 0.9 nesterov momentum and 0.0001 weight decay. The model is trained for 100 epochs with batch size 768. The initial learning rate is set to 0.3 and is reduced by 10 at epoch 30, 60, and 90. It takes $\sim$30 hours to train on 16 TESLA V100 GPUs.

\subsection{Details of Training \supernet on Semantic Segmentation}
\label{semantic_seg_training_details}

\textbf{Teacher training:} For a fair comparison, we follow the same training protocols in~\cite{wang2020deep}. We adopt the SGD optimizer with the momentum of 0.9 and the weight decay of 0.0005. The model is trained for 484 epochs with the batch size of 24 on 8 TESLA V100 GPUs. The initial learning rate is set to 0.01 and the cosine annual decay~\cite{loshchilov2016sgdr} is used for decaying the learning rate. For data preprocessing, the training and validation image size is 512×1024 and 1024×2048, respectively. For data augmentation strategies, we use random cropping (from 1024×2048 to 512×1024), random scaling (between [0.5, 2]), and random horizontal flipping.

\textbf{SuperScaleNet-Seg training:} We follow the same training protocols as teacher training except the initial learning rate is set to 0.001. 
This is because the SuperScaleNet-Seg is initialized from the well-trained teacher, we only need to fine-tune each sub-network using a small learning rate.

It takes $\sim$40(60) hours to obtain \supernet-Seg-W32(W48), including the teacher training. With only twice the training cost as stand-alone model training, we can obtain a series of segmentation models in a wide spectrum of FLOPs without additional retraining. We further use multi-scale topology evolution to explore elite \netname-Seg. 

\subsection{Details of Training \supernet on Top-Down Human Pose Estimation}

\textbf{Teacher training:} Following the training protocols of HRNet~\cite{wang2020deep}, we train the model for 210 epochs using the Adam optimizer~\cite{kingma2014adam} with step learning rate decay~\cite{xiao2018simple, wang2020deep}. The initial learning rate is set as 0.001, and is dropped to 0.0001 and 0.00001 at the 170th and 200th epochs, respectively. For data preprocessing, we extend the human detection box in height or width to a fixed aspect ratio -- height : width = 4 : 3, and then crop the box from the image, which is resized to a fixed size, 256 × 192 or 384 × 288. For data augmentation strategies, we use random rotation ([-45$^{\circ}$, 45$^{\circ}$]), random scale ([0.65, 1.35]), and flipping.

\textbf{SuperScaleNet-Pose training:} We follow the same training protocols in teacher training. We do not reduce the learning rate as in SuperScaleNet-Seg training because the Adam optimizer can adjust the learning rate adaptively~\cite{kingma2014adam}. 

The models are trained on 8 TESLA V100 GPUs. It takes $\sim$50(75) hours to train \supernet-Pose-W32(W48), including the teacher training. After topology evolution, we further fine-tune the \netname-P for 20 epochs (around 3 hours) for better performance.

\textbf{MPII:} We use the same data augmentation and training strategy for MPII, except that the input size is cropped to 256 × 256 for a fair comparison with SimpleBaseline~\cite{xiao2018simple} and HRNet~\cite{sun2019deep}.

\subsection{Details of Training found architectures on Bottom-Up Human Pose Estimation}
We train \netname-P series on bottom-up human pose estimation framework, HigherHRNet~\cite{cheng2020higherhrnet}. For a fair comparison, we use the exact same training routine as HigherHRNet. Specifically, we train our model for 300 epochs using the Adam optimizer~\cite{kingma2014adam} with step learning rate decay. The base learning rate is set to 0.001, and dropped to 1e-4 and 1e-5 at the 200th and 260th epochs, respectively. For data augmentation strategies, we use random rotation ([-30$^{\circ}$, 30$^{\circ}$]), random scale ([0.75, 1.5]), random translation ([-40, 40]), random crop (512 × 512), and random flip. We use the top-down SuperScaleNet-Pose to initialize weights.

\begin{figure*}
    \centering
    \includegraphics[width=\linewidth]{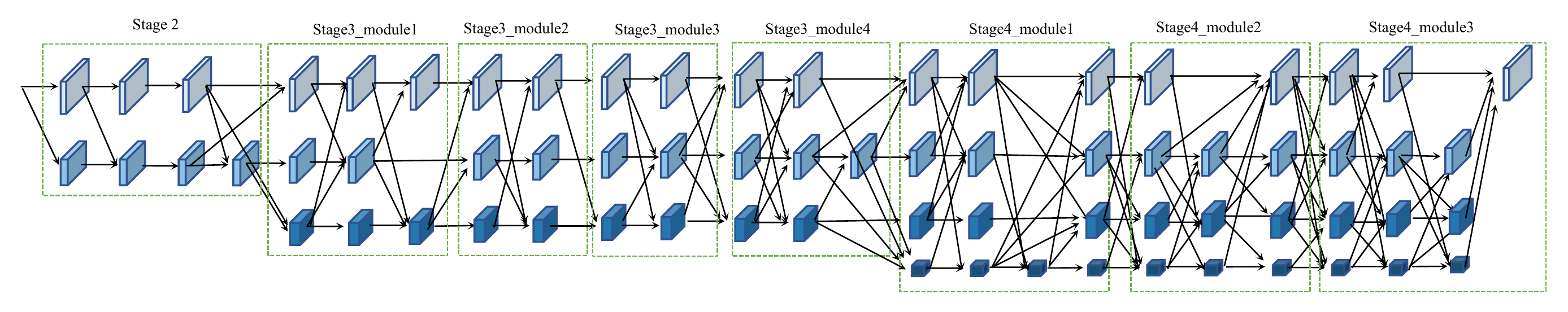}
    \caption{The full model of \netname-S1.}
    \label{fig:full_s1}
\end{figure*}
\begin{figure*}
    \centering
    \includegraphics[width=\linewidth]{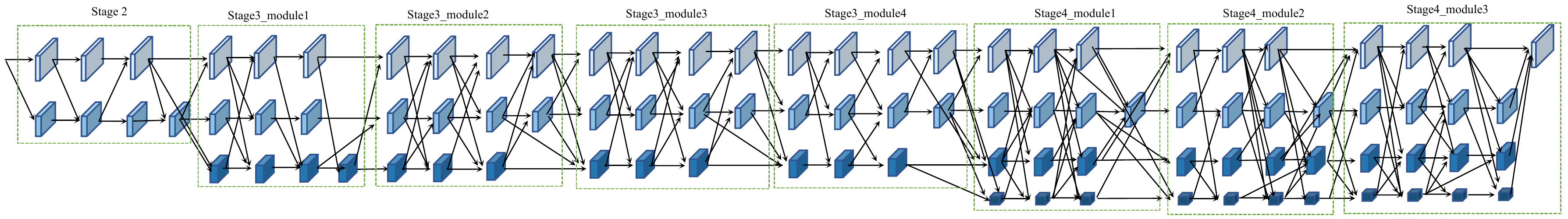}
    \caption{The full model of \netname-P1.}
    \label{fig:full_p1}
\end{figure*}

\subsection{Results on Object Detection}

We directly apply the \netname-S2, which is obtained from semantic segmentation, to object detection task. 
We plug in the \netname-S2 to two classic object detection frameworks, Faster R-CNN~\cite{ren2015faster} and Mask R-CNN~\cite{he2017mask} as shown in Table~\ref{tab:object_detection}. We use the whole COCO \textit{trainval135} as training set and validate on COCO \textit{minival}. For both Faster R-CNN and Mask R-CNN, the input images are resized to a short side of 800 pixels and a long side not exceeding 1333 pixels. We use SGD as optimizer with 0.9 momentum. For a fair comparison, all our models are trained for 12 epochs, known as 1$\times$ scheduler. We use 8 TESLA V100 GPUs for training with 16 global batch size. The initial learning rate is 0.02 and is divided by 10 at 8 and 11 epochs. 

In the Faster R-CNN framework, our networks perform better than HRNet-w32 with less parameters and computation cost. Our \netname-S2 is especially effective for small objects (1.1\% improvement for AP$_{S}$). The reason is that our \netname-S2 learns more high-resolution features which are beneficial for small objects. 

\subsection{Details of Found Architecture}
Here we provide the architecture structures of our crafted architecture \netname-P1 for human pose estimation and \netname-S1 for semantic segmentation, see Figure~\ref{fig:full_s1} and Figure~\ref{fig:full_p1}, respectively. 
The interesting observation is that both architectures have more multi-scale feature fusion at later stages while with a relatively simple network structure at the early stages.

\subsection{NAS Reproducibility Checklist}
To ensure a fair comparison, we follow the guidelines provided by NAS reproducibility checklist~\cite{lindauer2019best} and compare \workname with other NAS methods from different perspectives.

\begin{todolist}
    \item[\done] \textit{For all NAS methods you compare, did you use exactly the same NAS benchmark, including the same dataset (with the same training-test split), search space and code for training the architectures and hyperparameters for that code?}
    \begin{itemize}
        \item [--]  
        For comparing with other NAS methods~\cite{gong2020autopose,li2020learning}, we used the same dataset including train-test split. Our search space is essentially different from previous works. To train our architecture, we used the open-sourced repository from HRNet and HigherHRNet with the only change of learning rate. 
    \end{itemize}

    \item[\done] \textit{Did you control for confounding factors (different hardware, versions of DL libraries, different runtimes for the different methods)?} 
    \begin{itemize}
        \item [--] Yes, for the version of DL libraries, we used Pytorch-1.1 for conducting all our experiments and collecting our results. All the package dependencies are described in \texttt{requirements.txt} of our attached codes. For hardware, we only trained and tested on NVIDIA TESLA V100 GPU. 
    \end{itemize}
    
    \item[\done] \textit{Did you run ablation studies?} 
    \begin{itemize}
        \item [--]     Yes, we performed ablation studies for sampling method, searching method, and knowledge distillation. Detailed results can be found in Section 4.3 of our main submission.
        \end{itemize}
        
    \item[\done] \textit{Did you use the same evaluation protocol for the methods being compared?}
    \begin{itemize}
        \item [--] Yes, for evaluating on top-down human pose estimation, we followed the same evaluation protocol as HRNet paper. For comparing with bottom-up human pose estimation, we used the same evaluation protocol as HigherHRNet. For comparing on semantic segmentation, we used the same evaluation protocol as HRNet as well. 
    \end{itemize}
    \item[\done] \textit{Did you compare to random search?}
    \begin{itemize}
        \item  [--] Yes, we compared our searching method with random search at Section 4.3 and Figure 7 of our main submission. The results show that the Pareto front of our found architectures performs better than random search. 
    \end{itemize}

    \item[\done] \textit{Did you perform multiple runs of your experiments?} 
    \begin{itemize}
        \item [--]  Yes, during search space exploration, we trained the original HRNet 5 times and reported the mean and standard deviation of mIoU on Cityscapes. For  \supernet, we only trained once. We expect further tuning and training should achieve better results. 
    All of our experiment is highly reproducible and source code is provided.
    \end{itemize}

    \item \textit{Did you use tabular or surrogate benchmarks for in-depth evaluations?} 
        \begin{itemize}
            \item [--] No, existing surrogate benchmarks such as NASBench-101, NAS-Bench-201, NAS-Bench-1Shot1 are not application to our search space.
        \end{itemize}
    \item[\done] \textit{Did you report how you tuned hyperparameters, and what time and resources this required?} 
    \begin{itemize}
        \item [--] Yes, among all the hyperparameters, we only tuned the learning rate for training \supernet on semantic segmentation tasks. Since \supernet-Seg is initialized from the well-trained teacher, the original learning rate (0.01) used in HRNet is not suitable for our case. We tried three different learning rates, 0.005, 0.002, and 0.001. Each tuning cost 300 GPU hours, with a total cost of around 900 GPU hours on TESLA V100. 
        We found that the learning rate of 0.001 shows the best performance, thus, we use 0.001 for semantic segmentation tasks.
    \end{itemize}

    \item[\done] \textit{Did you report all the details of your experimental setup?} 
    \begin{itemize}
        \item [--] Yes, we comprehensively reported all of the configurations, including hyperparameter settings, training protocol in main submission and supplementary material. 
    \end{itemize}
    
\end{todolist}